\documentclass[runningheads]{llncs}
\usepackage[T1]{fontenc}
\usepackage{times}
\usepackage{soul}
\usepackage{url}
\usepackage[hidelinks]{hyperref}
\usepackage[utf8]{inputenc}
\usepackage[small]{caption}
\usepackage{graphicx}
\usepackage{subcaption}
\usepackage{amsmath}
\usepackage{booktabs}
\usepackage{algorithm}
\usepackage{algorithmic}
\usepackage{amssymb}
\usepackage{pifont}
\usepackage{times}
\usepackage{epsfig}
\usepackage{amssymb}
\usepackage{amsfonts}
\usepackage{algorithmic}
\usepackage{textcomp}
\usepackage{threeparttable}
\usepackage{multicol}
\usepackage{multirow}
\usepackage{xspace}
\usepackage{bm}
\usepackage{balance}
\usepackage{float}
\usepackage{color}
\usepackage{ragged2e}
\usepackage{mathrsfs}
\def\ie{\emph{i.e.,~}}
\def\eg{\emph{e.g.,~}}
\usepackage{color}
\def\percepbench{MorphoPercept-Bench}
\def\classibench{CytoCell-Bench}
\def\ournet{Singpath-VL}
\begin{document}

\definecolor{qiu}{RGB}{250,10,10}
\def\qiu{\textcolor{qiu}}

\title{Singpath-VL Technical Report}
%


\author{Zhen Qiu, Kaiwen Xiao, Zhengwei Lu, Xiangyu Liu, Lei Zhao, Hao Zhang}  
\authorrunning{Zhen Qiu et al.}
\institute{LBP Singpath AI Lab}
  
\maketitle              
\begin{abstract}
We present \textbf{\ournet}, a vision-language large model, to fill the vacancy of AI assistant in cervical cytology. Recent advances in multi-modal large language models (MLLMs) have significantly propelled the field of computational pathology. However, their application in cytopathology, particularly cervical cytology, remains underexplored, primarily due to the scarcity of large-scale, high-quality annotated datasets. To bridge this gap, we first develop a novel three-stage pipeline to synthesize a million-scale image-description dataset. The pipeline leverages multiple general-purpose MLLMs as weak annotators, refines their outputs through consensus fusion and expert knowledge injection, and produces high-fidelity descriptions of cell morphology. Using this dataset, we then fine-tune the Qwen3-VL-4B model via a multi-stage strategy to create a specialized cytopathology MLLM. The resulting model, named \ournet, demonstrates superior performance in fine-grained morphological perception and cell-level diagnostic classification. To advance the field, we will open-source a portion of the synthetic dataset and benchmark.


\end{abstract}

\section{Introduction}

The advent of multimodal large language models (MLLMs) has marked a paradigm shift in artificial intelligence, enabling unified understanding and generation across vision, language, and other modalities. General-purpose MLLMs, pre-trained on vast and diverse corpora, have demonstrated remarkable zero-shot capabilities in describing medical images and answering related questions. For computational pathology, this technology holds transformative potential, offering avenues for automated diagnostic assistance, comprehensive slide analysis, and interactive educational tools.

Despite this progress, the application of MLLMs in the specialized sub-field of \textit{cytopathology}---particularly in cervical cytology screening based on The Bethesda System (TBS)---remains significantly underexplored. A primary and formidable bottleneck is the acute scarcity of large-scale, high-quality, publicly available datasets that pair cytopathology images with detailed, expert-level morphological descriptions. Unlike histopathology, where tissue architecture provides rich, contextual visual information, cytopathology diagnosis relies on the precise interpretation of subtle cellular and nuclear features (e.g., nuclear hyperchromasia, irregular nuclear membranes, and increased nuclear-to-cytoplasmic ratio) from individual cells or small cell clusters. This demands a model with an exceptionally fine-grained visual grounding capability, for which generic medical image-text data is insufficient.

Current approaches often rely on manually curated, small-scale datasets or adapt general vision-language models with limited success, failing to capture the nuanced expertise required for accurate cytodiagnosis. While some studies have leveraged foundation models as weak annotators to generate data in other domains, a tailored, robust pipeline for the high-stakes, detail-sensitive domain of cytopathology is lacking. The need, therefore, is twofold: (1) a scalable method to generate faithful and rich cytology-specific image-text data, and (2) a specialized model trained effectively on such data to master the domain's visual lexicon and diagnostic logic.

To address these challenges, we present \textbf{\ournet}, a novel vision-language assistant for cervical cytology. Our work makes three key contributions:

\begin{itemize}
    \item \textbf{Novel Data Generation Pipeline:} We introduce an automated, three-stage pipeline to synthesize a million-scale, high-fidelity image-description dataset (\textit{Singpath-CytoText}) from an internal repository of cervical cytology image tiles. This pipeline strategically employs multiple open-source MLLMs (e.g., Qwen3-VL, InternVL-3.5) as \textit{weak annotators}, fuses their consensus descriptions via a large language model, and finally refines the output with a domain-specific expert model to ensure accuracy and completeness.

    \item \textbf{Specialized Model Training:} We develop \ournet~by fine-tuning the Qwen3-VL-4B foundation model on our generated dataset through a multi-stage training recipe. This regimen includes vision-language alignment, supervised fine-tuning for instruction following, and a knowledge replay strategy to mitigate catastrophic forgetting, thereby balancing deep domain adaptation with the retention of beneficial general knowledge.

    \item \textbf{Superior Performance \& Open-Source Contribution:} We demonstrate that the proposed method significantly outperforms baseline general-purpose MLLMs in the perception of fine-grained morphological patterns and in cell-level diagnostic classification tasks. To promote further research and transparency, we commit to open-sourcing a portion of the \textit{Singpath-CytoText} dataset.
\end{itemize}

The remainder of this paper is organized as follows: Section~\ref{method} details our data generation pipeline and model training strategy. Section~\ref{exp} presents experimental results and analysis. Finally, Section~\ref{future} discusses future directions.

\section{Method}
\label{method}
In this section, we detail the proposed data curation pipeline and training recipe.
\subsection{Data Curation}

The development of a specialized vision-language model for cervical cytology is fundamentally constrained by the scarcity of large-scale, high-quality annotated datasets. Given that general-purpose MLLMs have been trained on some cytological content, we exploit them to generate image description data. We begin by evaluating a suite of the latest open-source MLLMs, including Qwen3-VL-32B, InternVL-3.5-38B, Baichuan-Omni-1.5, and Medgemma-27B. Our preliminary study reveals significant disparities in their ability to accurately identify different morphological features (\eg nuclear-to-cytoplasmic ratio, chromatin pattern). To this end, we propose to leverage the complementary strengths of these state-of-the-art, open-source MLLMs as \textit{weak annotators} to synthesize caption for image tiles. 

To synthesize a large-scale high-fidelity dataset, we designed a novel three-stage generation pipeline. This pipeline intelligently aggregates and refines outputs from multiple foundation models, followed by targeted expert refinement.

\subsubsection{Stage 1: Parallel Image Caption Generation}
Given an image tile cropped from a cervical cytology WSI, we query three distinct, powerful MLLMs, \ie Qwen3-VL-32B, InternVL-3.5-38B, and Baichuan-Omni-1.5, in parallel. Each model is prompted to generate a textual description focused on a predefined set of key morphological analysis dimensions relevant to TBS-based cell classification. This yields three independent, and potentially diverse, preliminary descriptions.

\subsubsection{Stage 2: Consensus Fusion and Coherent Summarization}
The three descriptions from Stage 1 are fed into a large language model (LLM) acting as an \textit{information integrator}. The LLM is instructed to perform two key tasks:
\begin{itemize}
    \item \textbf{Consensus Extraction and Conflict Resolution:} Identify morphological features consistently described across the majority of the input texts.
    \item \textbf{Feature Filtering and Text Rewriting:} Automatically filter out feature points that the foundation models consistently fail to perceive correctly due to a lack of model capability, deeming them \textit{model-missing dimensions}. The LLM then synthesizes the consensus information into a single, fluent, and structured paragraph.
\end{itemize}

\subsubsection{Stage 3: Expert Knowledge Injection}
To address the \textit{model-missing dimensions} filtered out in Stage 2 and to inject domain-specific precision, we introduce a final refinement stage. A smaller, finely-tuned \textit{expert model}—specialized in cervical cytopathology—analyzes the original image patch and the fused description of the Stage 2. Its role is to supplement the description with any critical, fine-grained morphological details that the generalist MLLMs may have overlooked or described inadequately, producing the final, high-quality description.

\subsubsection{Generated Dataset}
By applying this pipeline to our extensive internal repository of cervical cytology image tiles, we successfully constructed \textbf{Singpath-CytoText}, a million-scale multimodal dataset. Each data point consists of an image tile paired with a description that combines structured morphological insights with coherent narrative summary text, providing a robust foundation for training sophisticated cytopathology-specific MLLMs.

\subsection{Training Recipe}
\begin{figure}[htbp]
    \centering
    \includegraphics[width=\linewidth]{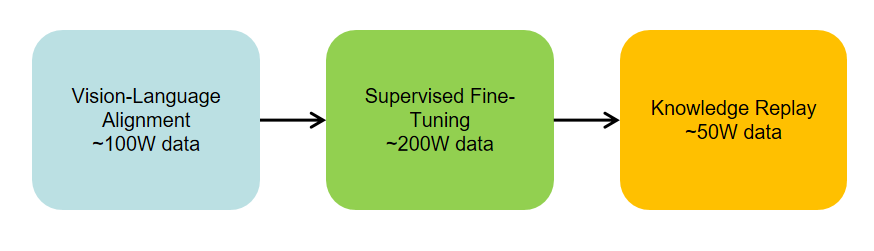}
    \caption{Overview of the multi-stage training recipe for \ournet. The pipeline consists of three consecutive stages: (1) Vision-Language Alignment for domain grounding, (2) Supervised Fine-Tuning for instruction following, and (3) Knowledge Replay for mitigating catastrophic forgetting, resulting in a specialized and robust cervical cytology MLLM.}
    \label{fig:pipeline}
\end{figure}

To develop a multimodal large language model (MLLM) for cytopathology, we employ the latest Qwen3-VL-4B as our base model. Our fine-tuning strategy consists of three sequential stages designed to endow the model with domain-specific visual understanding, instruction-following capability, while mitigating catastrophic forgetting. The overall pipeline is illustrated in Figure~\ref{fig:pipeline}.

\subsubsection{Stage 1: Vision-Language Alignment}
\label{sec:stage1}
To align the visual encoder with the language decoder within the cytopathological domain, enabling the model to generate accurate and relevant textual descriptions from cellular images.
We perform full-parameter fine-tuning on the proposed \textit{Singpath-CytoText}. This stage focuses on establishing fundamental grounding between visual features (\eg cell morphology, staining patterns) and their corresponding pathological terminology.

\subsubsection{Stage 2: Supervised Fine-Tuning for Instruction Following}
\label{sec:stage2}
To enhance the model's ability to follow complex, structured instructions specific to cytopathological diagnosis and reporting. The image-description pairs from Stage~1 are reformatted into an instruction-following dataset. This transformation is guided by structured cytopathology observation templates (\eg reporting elements from The Bethesda System). The resulting dataset contains both single-turn and multi-turn dialogues, simulating real-world diagnostic queries and iterative analysis.

\subsubsection{Stage 3: Knowledge Replay for Mitigating Forgetting}
\label{sec:stage3}
To counteract catastrophic forgetting of both domain-specific and general knowledge during the intensive fine-tuning stages. We employ a knowledge replay strategy using two distinct data streams:
\begin{itemize}
    \item \textbf{Domain-Specific Knowledge:} We utilize the cytopathology data from previous stages and generate additional pure-text question-answering (QA) pairs using the original Qwen3-VL 4B.
    \item \textbf{General-Domain Knowledge:} We sample a subset of general-domain image and similarly generate corresponding VQA pairs using the original Qwen3-VL 4B.
\end{itemize}
\textbf{Rationale for Using the Original Model:} The purpose of employing the base model (Qwen3-VL 4B), rather than a more powerful one, for data generation is to perform model regularization. This approach gently guides the fine-tuned model, helping to preserve its original reasoning framework and reduce excessive deviation from the foundation model's broad capabilities.

\section{Evaluation}
\label{exp}
\subsection{Benchmark Construction}
In the clinical workflow of liquid-based cervical cytology, pathologists make diagnostic decisions based on detailed morphological features of cells. To rigorously evaluate our model's ability to perceive and interpret these fine-grained morphological characteristics, we constructed a specialized benchmark dataset for morphological perception.

A panel of experienced cytopathologists performed detailed, fine-grained annotations on a collection of cell images. Specifically, they provided binary classification labels (e.g., present/absent, normal/abnormal) for nine key morphological observations critical to diagnosis, such as nuclear size, chromatin pattern, and nuclear-to-cytoplasmic ratio. This process resulted in \textbf{\percepbench}, a high-quality dataset designed to quantitatively assess a model's morphological perception accuracy against expert-established ground truth. The data statistic is shown in Table~\ref{tab:interpre}

To further evaluate the model's end-to-end diagnostic capability within the standardized reporting framework, we constructed a large-scale \textbf{\classibench}. This dataset comprises 29102 cell images, each annotated by certified cytopathologists with its corresponding diagnostic category according to \textit{The Bethesda System (TBS)}. The diagnostic categories include NILM, ASC-US, LSIL, ASC-H, HSIL and AGC. The data statistic is shown in Table~\ref{tab:cellsta}.

\begin{table*}[!hbt]
\setlength\tabcolsep{3pt}
    \begin{center}
    \scalebox{1.0}{
         \begin{tabular}{lcccccccccc}
         \toprule
          & NE & NA & NH & Koilocyte & CT & Nucleolus & NC & NCR & NM & Total  \\
         \midrule
         Num. & 153 & 151 & 146 & 393 & 131 & 360 & 396 & 231 & 108 & 2069 \\
        \bottomrule
        \end{tabular}
    }
    \end{center}
    \caption{\label{tab:interpre}The data statistic of \percepbench. Note that NE, NA, NH, CT, NC, NCR, NM denote Nuclear Enlargement, Nuclear Atypia, Nuclear Hyperchromasia, Chromatin Texture, Nuclear Count, Nuclear-to-Cytoplasmic Ratio, Nuclear Membrane, respectively.}
\end{table*}

\begin{table*}[!hbt]
\setlength\tabcolsep{3pt}
    \begin{center}
    \scalebox{1.0}{
         \begin{tabular}{lccccccc}
         \toprule
          & NILM & ASC-US & LSIL & ASC-H & HSIL & AGC & Total  \\
         \midrule
         Num. & 10413 & 4157 & 2340 & 5987 & 4786 & 1419 & 29102 \\
        \bottomrule
        \end{tabular}
    }
    \end{center}
    \caption{\label{tab:cellsta}The data statistic of \textbf{\classibench}.}
\end{table*}

\subsection{Morphological Perception and Recognition}

\begin{figure}[htbp]
    \centering
    \includegraphics[width=0.8\linewidth]{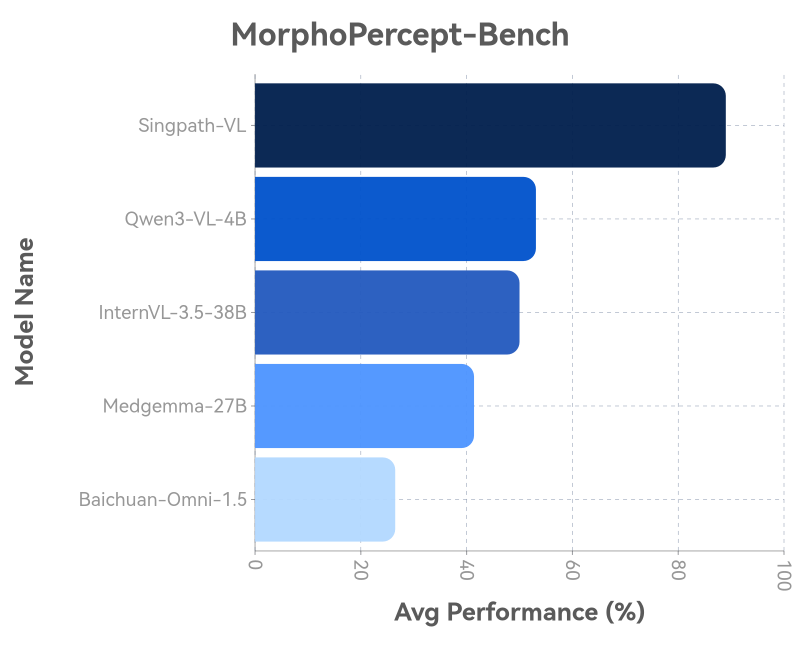} 
    \caption{Comparison of average performance scores (\%) on \percepbench. This bar chart aggregates the mean values across all key morphological observations.}
    \label{fig:perc_bench}
\end{figure}

We evaluated the performance of our proposed model against several leading open-source multi-modal large language models (MLLMs) on \percepbench. The results, summarized in Table~\ref{tab:percep} and Figure~\ref{fig:perc_bench}, demonstrate that our model achieves a significant and substantial improvement over all compared baselines. Note that all compared baselines conduct zero-shot inference on \percepbench.

Notably, our model's classification accuracy on this morphological perception task not only surpasses other models but also exceeds the average inter-rater agreement level among the expert annotators themselves. This indicates that our model has learned a robust and precise representation of cytomorphological features, enabling it to perform at a level comparable to or exceeding the consistency of human experts in this focused perceptual task.

\begin{table*}[!hbt]
\setlength\tabcolsep{3pt}
    \begin{center}
    \scalebox{0.95}{
         \begin{tabular}{lcccccccccc}
         \toprule
         Method & NE & NA & NH& Koilocyte & CT & Nucleolus & NC & NCR & NM & Avg \\
         \midrule
         Inter-rater agreement & 72.4 & 72.2 & 72.3 & 96.2 & 63.6 & 91.4 & 95.8 & 81.4 & 58.6 & 78.2 \\
         \midrule
         Qwen3-VL-4B~\cite{Qwen3-VL} & 44.5 & 66.7 & 37.9 & 65.2 & 19.9 & 73.9 & 60.6 & 53.6 & 55.8 & 53.1 \\
         InternVL-3.5-38B~\cite{wang2025internvl3_5}  & 43.1 & 40.5 & 45.1 & 17.9 & 68.0 & 27.5 & 77.3 & 58.0 & 72.7 & 50.0 \\
         Medgemma-27B~\cite{sellergren2025medgemma}  & 42.8 & 39.9 & 37.9 & 19.9 & 46.2 & 65.9 & 57.9 & 51.3 & 11.0 & 41.4 \\
         Baichuan-Omni-1.5~\cite{li2025baichuan} & 40.8 & 37.6 & 41.0 & 15.2 & 11.2 & 2.9 & 32.7 & 52.7 & 4.5 & 26.5 \\
        \midrule
        \ournet  & \textbf{91.2} & \textbf{95.7} & \textbf{90.8} & \textbf{96.1} & \textbf{75.2} & \textbf{94.1} & \textbf{98.0} & \textbf{78.7} & \textbf{80.8} & \textbf{89.0} \\
        \bottomrule
        \end{tabular}
    }
    \end{center}
    \vspace{-0.1in}
    \caption{\label{tab:percep}Classification accuracies (\%) on \textbf{\percepbench}. Note that NE, NA, NH, CT, NC, NCR, NM denote Nuclear Enlargement, Nuclear Atypia, Nuclear Hyperchromasia, Chromatin Texture, Nuclear Count, Nuclear-to-Cytoplasmic Ratio, Nuclear Membrane, respectively.}
\end{table*}

\begin{table*}[!hbt]
\setlength\tabcolsep{3pt}
    \begin{center}
    \scalebox{1.0}{
         \begin{tabular}{lccccccc}
         \toprule
         Method & NILM & ASC-US & LSIL & ASC-H & HSIL & AGC & Avg \\
         \midrule
         EfficientNet B0~\cite{tan2019efficientnet} & 100.0 & 69.6 & 73.1 & 73.5 & 79.4 & 89.3 & 80.8 \\
         \midrule
         Qwen3-VL-4B~\cite{Qwen3-VL} & 0.0 & 0.2 & 0.0 & 99.7 & 0.0 & 0.0 & 16.7 \\
         InternVL-3.5-38B~\cite{wang2025internvl3_5}  & 7.3 & 2.7 & 25.5 & 4.2 & 61.3 & 8.7 & 18.3 \\
         Medgemma-27B~\cite{sellergren2025medgemma}  & 0.0 & 0.0 & 0.0 & 0.0 & 100.0 & 0.0 & 16.7 \\
         Baichuan-Omni-1.5~\cite{li2025baichuan} & 0.1 & 3.2 & 0.1 & 66.0 & 22.1 & 7.1 & 16.4 \\
        \midrule
        \ournet  & \textbf{100.0} & \textbf{75.1} & \textbf{71.8} & 77.2 & 73.5 & \textbf{83.8} & \textbf{80.2} \\
        \bottomrule
        \end{tabular}
    }
    \end{center}
    \vspace{-0.1in}
    \caption{\label{tab:cellclassi}Classification accuracies (\%) on \textbf{\classibench}.}
\end{table*}

\begin{figure}[htbp]
    \centering
    \includegraphics[width=0.8\linewidth]{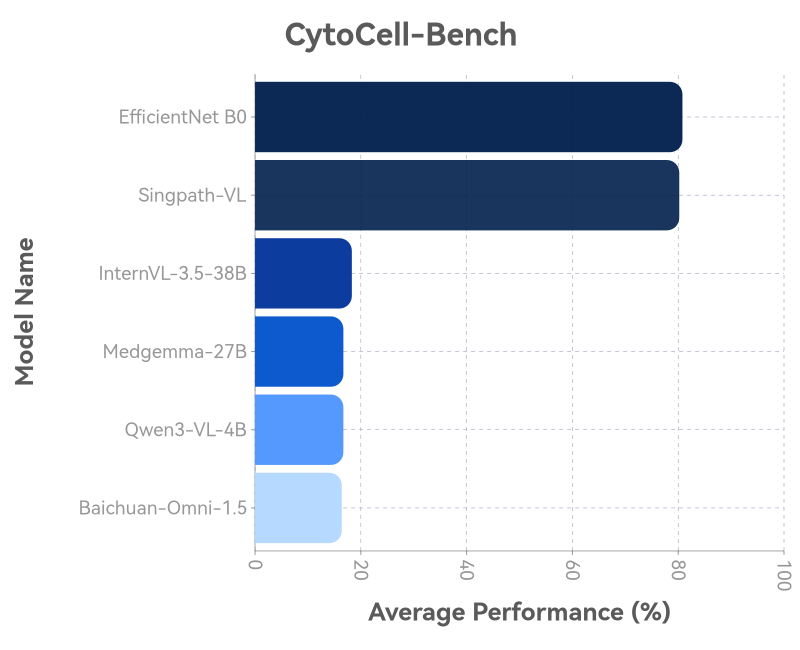} 
    \caption{Comparison of average performance scores (\%) on \classibench. This bar chart aggregates the mean values across 6 diagnostic categories.}
    \label{fig:classi_bench}
\end{figure}

\begin{figure}[t]
    \centering
    \includegraphics[width=\linewidth]{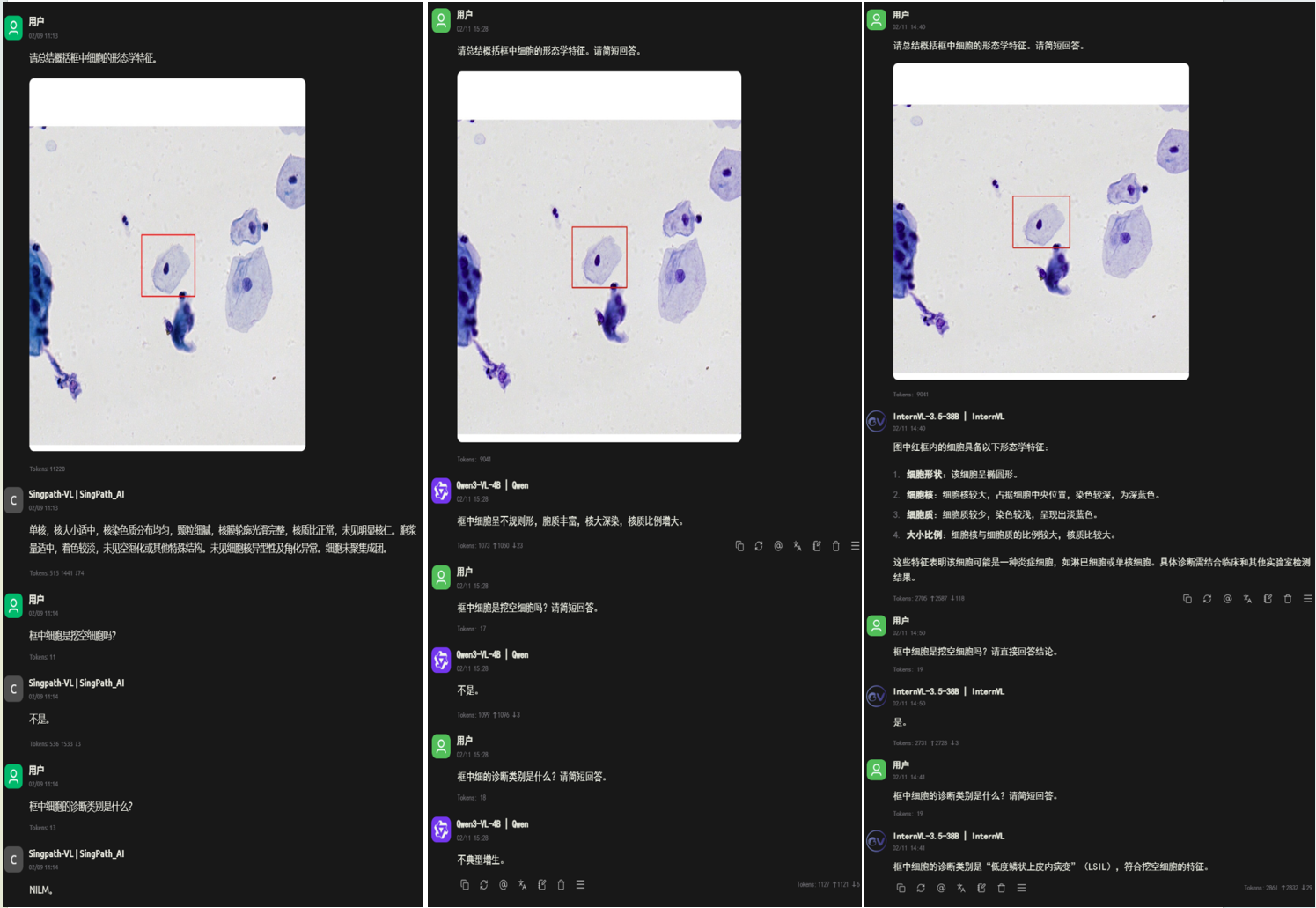}
    \caption{Qualitative comparison of responses from three vision-language models (\ournet, Qwen3-VL-4B, and InternVL-3.5-38B from left to right) on a sample of Negative for Intraepithelial Lesion or Malignancy (NILM).}
    \label{fig:quali_result}
\end{figure}

\begin{figure}[t]
    \centering
    \includegraphics[width=\linewidth]{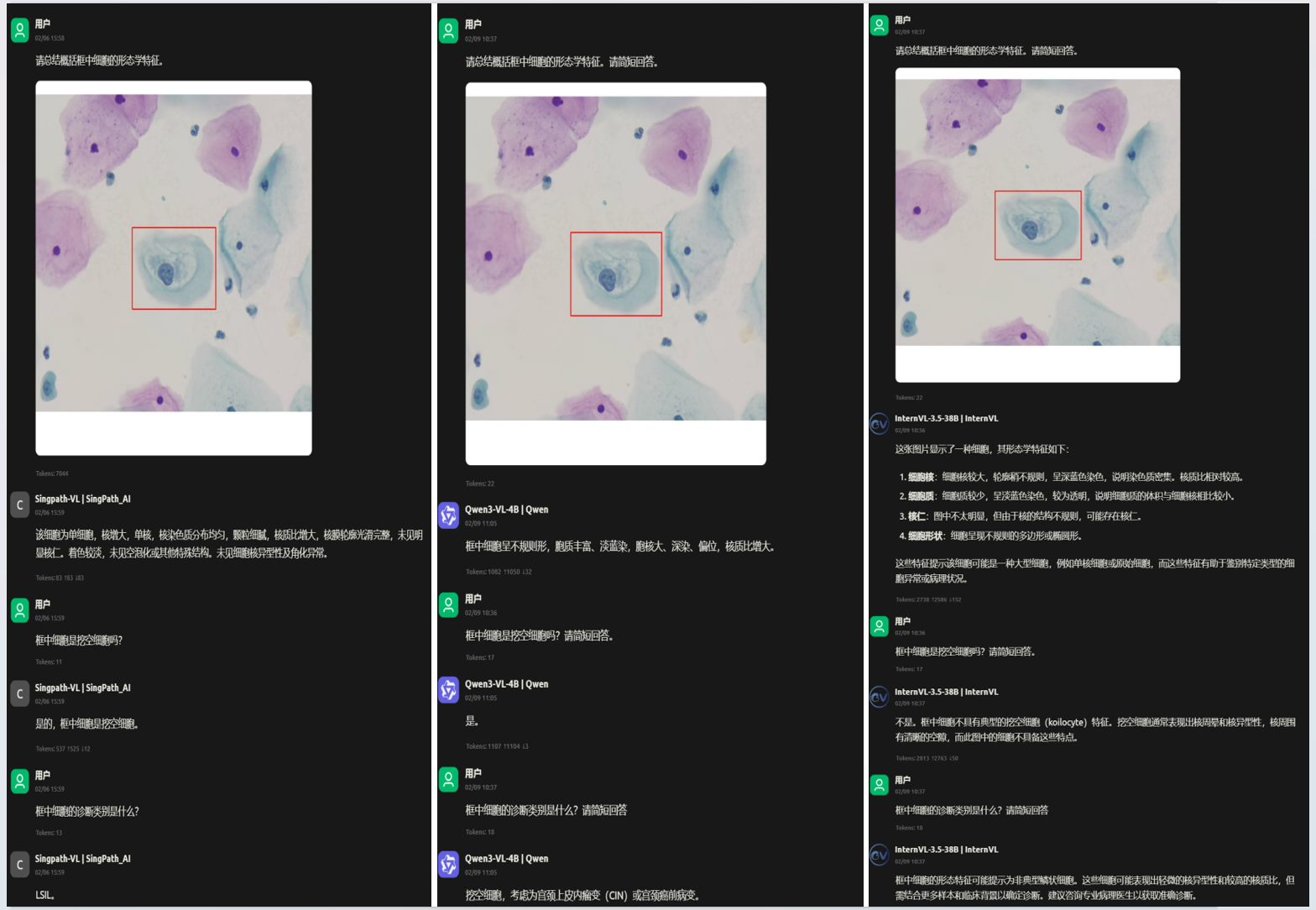}
    \caption{Qualitative comparison of responses from three vision-language models (\ournet, Qwen3-VL-4B, and InternVL-3.5-38B from left to right) on a sample of Low-Grade Squamous Intraepithelial Lesion (LSIL).}
    \label{fig:quali_result2}
\end{figure}

\subsection{TBS-based Cervical Cytology Classification}
Based on \classibench, we conduct evaluation on our model and several baselines. The results are shown in Table~\ref{tab:cellclassi} and Figure~\ref{fig:classi_bench}. Note that EfficientNet B0 is trained on the same dataset as our mllm model, while the other mllm baselines are tested through zero-shot inference. Our proposed model demonstrates competitive performance across all diagnostic categories of \classibench. Specifically, our model achieves perfect accuracy (100.0\%) on the NILM category, matching the performance of the EfficientNet B0 baseline. For the clinically challenging ASC-US and ASC-H categories--which represent ambiguous cellular changes--our model outperforms EfficientNet B0 by +5.5\% and +3.7\%, respectively, indicating its enhanced capability to discern subtle morphological distinctions. For general-purpose MLLMs, their predictions show a preferential tendency toward a particular class. As a result, they catastrophically fail on this task. Overall, our model achieves a balanced performance profile, excelling particularly in the diagnostically critical ambiguous categories.

\subsection{Qualitative Evaluation}
We qualitatively evaluate the morphological perception and cervical cytology classification capabilities of different vision-language models on cervical cytology images using two representative cases: NILM and LSIL. As shown in Figure~\ref{fig:quali_result} and Figure~\ref{fig:quali_result2}, only our specialized model, \ournet, provided morphologically accurate descriptions and the correct TBS classification for both cases. In contrast, the general-purpose models, Qwen3-VL-4B and InternVL-3.5-38B, exhibited errors in perceiving morphological characteristics, leading to incorrect diagnoses. Notably, these models exhibited a clear tendency to over-classify the samples as positive, likely due to biases in their general-purpose training data and a lack of nuanced domain-specific understanding required for precise cytopathological discrimination.

\section{Conclusion}
\label{future}
In this report, we introduce \textbf{\ournet}, a specialized vision-language large model for cervical cytology. Based on the million-scale image description dataset generated by a three-stage pipeline, we finetune the Qwen3-VL-4B through a three-stage training recipe. The resulting model demonstrates superior performance in perceiving fine-grained morphological features and performing cell-level diagnostic classification according to The Bethesda System (TBS).

\section{Limitations and Future}
\label{sec:limitations}

Despite the superior performance, our work has several limitations that point to directions for future improvement.

\textbf{Scope Specificity:} The current model is specifically designed and evaluated on \textbf{cervical cytology}. While this focus allows for depth, its architecture and training may not directly generalize to other cytopathology subspecialties (e.g., thyroid FNA, urine cytology) without substantial retraining and domain adaptation. The morphological criteria and diagnostic frameworks differ across these domains, necessitating targeted efforts for broader applicability.

\textbf{Limited Downstream Tasks:} Our current validation primarily focuses on morphological description generation and cell-level classification. While these are core tasks, the full potential of an AI assistant in a clinical workflow includes more complex downstream applications such as slide-level screening efficiency, integration with clinical history for risk assessment, and generating differential diagnoses. The model's performance on these integrated, higher-order tasks remains to be fully explored.

Further, we identify several promising avenues for future research.

\textbf{Towards Causal Reasoning and Diagnostic Inference:} The most significant direction is to move beyond pattern recognition and empower the model with explainable diagnostic reasoning capabilities. Instead of directly mapping an image to a TBS category, a future model could be trained to explicitly articulate the logical chain of inference: identifying key morphological features (e.g., "high nuclear-to-cytoplasmic ratio," "coarse chromatin") and then applying established diagnostic rules (e.g., "the presence of both features A and B suggests a high-grade squamous intraepithelial lesion, HSIL") to arrive at a conclusion. This would make the model's decision-making process transparent, auditable, and more aligned with the cytopathologist's cognitive process.

\textbf{Expansion to Pan-Cytopathology:} A natural extension is to broaden the model's scope to become a pan-cytopathology assistant. This involves curating datasets from various cytology subspecialties and developing mechanisms for the model to dynamically recognize and apply the appropriate diagnostic guideline (e.g., TBS for gynecologic cytology, the Paris System for urine cytology).

\textbf{Interactive and Human-in-the-Loop Systems:} Future iterations could evolve into interactive diagnostic support systems. The model could engage in multi-turn dialogue with pathologists, answering clarifying questions, highlighting ambiguous cells for review, and justifying its suggestions, thereby fostering a collaborative human-AI partnership.

\bibliographystyle{splncs04}
\bibliography{cytovler}

\end{document}